\documentclass{article}

\usepackage{microtype}
\usepackage{graphicx}
\usepackage{subcaption}
\usepackage{booktabs} 

\usepackage{hyperref}

\usepackage[preprint]{icml2026}

\usepackage{amsmath}
\usepackage{amssymb}
\usepackage{mathtools}
\usepackage{amsthm}

\usepackage{graphicx}
\usepackage{subcaption}
\usepackage{multirow}
\usepackage[normalem]{ulem}
\usepackage{listings}
\usepackage{bm}
\usepackage{pifont}
\usepackage{adjustbox}
\usepackage{hyperref}
\usepackage{float}
\usepackage{threeparttable}
\usepackage{algorithmic}
\usepackage{algorithm}
\usepackage{xcolor}
\usepackage{makecell}
\useunder{\uline}{\ul}{}

\renewcommand{\algorithmicrequire}{ \textbf{Input:}}     
\renewcommand{\algorithmicensure}{\textbf{Output:}}    

\usepackage[capitalize,noabbrev]{cleveref}

\theoremstyle{plain}

\theoremstyle{definition}

\theoremstyle{remark}

\usepackage[textsize=tiny]{todonotes}

\icmltitlerunning{AGoQ: Activation and Gradient Quantization for Memory-Efficient Distributed Training of LLMs}

\begin{document}
\def \modelname {AGoQ}
\twocolumn[
  \icmltitle{\modelname: Activation and Gradient Quantization for \\Memory-Efficient Distributed Training of LLMs}



  \icmlsetsymbol{equal}{*}

  \begin{icmlauthorlist}
    \icmlauthor{Wenxiang~Lin}{yyy}
    \icmlauthor{Juntao~Huang}{yyy}
    \icmlauthor{Luhan~Zhang}{yyy}
    \icmlauthor{Laiyi~Li}{yyy}
    \icmlauthor{Xiang~Bao}{comp}
    \icmlauthor{Mengyang~Zhang}{comp}
    \icmlauthor{Bing~Wang}{comp}
    \icmlauthor{Shaohuai~Shi}{yyy}
  \end{icmlauthorlist}

  \icmlaffiliation{yyy}{School of Computer Science and Technology, Harbin Institute of Technology, Shenzhen}
  \icmlaffiliation{comp}{Huawei Technologies Ltd}

  \icmlcorrespondingauthor{Shaohuai~Shi}{shaohuais@hit.edu.cn}

  \icmlkeywords{Machine Learning, Distributed training, Quantization, Memory-efficient}

  \vskip 0.3in
]



\printAffiliationsAndNotice{}  

\begin{abstract}
 Quantization is a key method for reducing the GPU memory requirement of training large language models (LLMs). Yet, current approaches are ineffective for 4-bit activations and 8-bit gradients, which would easily cause slow convergence or accuracy loss. To address this, we introduce \modelname, incorporating two new techniques: 1) a layer-aware activation quantization algorithm that allocates appropriate bit-widths for activations of various layers based on their types and pipeline stages to achieve near 4-bit activation storage, and 2) a gradient quantization algorithm that reduces memory usage and shortens communication time by employing 8-bit gradient storage and precision-preserving 8-bit All-Reduce communication. We conduct extensive experiments using different sizes of LLMs on two GPU clusters (up to 64 GPUs), and the experimental results show that our \modelname{} reduces the memory by up to 52\% and achieves up to 1.34$\times$ improvement of training speed compared to state-of-the-art training systems Megatron-LM (w/ or w/o ZeRO), COAT and DeepSpeed with 8B to 32B LLaMA models, while achieving convergence loss on pretraining and comparable accuracy on downstream tasks with LLaMA architectures.
\end{abstract}

\section{Introduction}
Distributed training has become a de-facto approach to accelerate the training process of deep neural networks (DNNs) on multi-GPU/TPU clusters~\cite{DBLP:dean2012large,jia2018highly,narayanan2021efficient}. Particularly, data parallelism (DP) has been widely used by distributing training data to different workers (or GPUs) to train a model collaboratively~\cite{DBLP:dean2012large}. However, with the model size significantly increased as seen in large language models (LLMs), the memory requirement for training LLMs becomes a significant pressure~\cite{DBLP:Brown2020GPT3}. Thus, training LLMs typically requires using model parallelism, including tensor parallelism (TP)~\cite{narayanan2021efficient} and pipeline parallelism (PP)~\cite{huang2019gpipe}, which partition model parameters across different devices so that each GPU has enough memory to store the required data. 
DP, TP, and PP have been default features in the popular LLM training framework Megatron-LM~\cite{narayanan2021efficient}. 
Another popular memory-efficient training system, DeepSpeed~\cite{rasley2020deepspeed} exploits the zero redundancy optimizer (ZeRO) series (ZeRO-1/2/3) ~\cite{rajbhandari2020zero,ren2021zerooffload,rajbhandari2021zero} to save memory in LLM training. The newly introduced fully sharded data parallelism (FSDP)~\cite{zhao2023pytorchfsdp} in PyTorch's ecosystem has a similar idea to ZeRO-3. 

The device memory occupation of training an LLM mainly consists of model parameters, gradients, optimizer states, and temporary activations. Among them, the activations typically occupy the largest proportion of memory and are linearly increased with the increase of sequence length and batch size, which are two common hyper-parameters. There have been extensive studies aimed at reducing the memory footprint by reducing the occupation of activations including activation recomputation or offloading~\cite{chen2016training,yuan2024accelerating,wu2025ssdtrain} and activation quantization~\cite{evans2021ac,liu2022gact,xi2024jetfire,xi2024coat,shamshoum2025compact,chen2025adacc}. Activation recomputation (or offloading) is a system-level optimization technique; thus, it has no side effects on model accuracy, but it introduces extra overhead by recomputing (or uploading) activations for backpropagation. Activation quantization, on the other hand, uses low-precision formats (e.g., INT8~\cite{xi2024jetfire} and FP8~\cite{xi2024coat}) to store activation values and dequantizes them back to BF16/FP16 for backpropagation. However, the quantization and dequantization processes (even when using only 8-bit INT8 or FP8) result in accuracy loss compared to pure BF16/FP16~\cite{xi2024jetfire,xi2024coat}. Jetfire~\cite{xi2024jetfire} and COAT~\cite{xi2024coat} attempt to address the accuracy loss problem of 8-bit activation quantization by employing dynamic quantization and block-wise quantization, \textit{but they are still not applicable to lower-bit formats (e.g., 4-bit)}. 

Furthermore, in terms of gradient memory, although there are extensive works~\cite{tang20211,bai2021gradient,shi2021towards,peng2023birder,huang2024gzccl,wang2024zeroplusplus} trying to compress gradients to reduce communication overheads, the gradients are still stored in high-precision (FP32) and quantized to low-precision for communication, which means the gradients still occupy the same size of memory as model parameters. One notable study by Microsoft~\cite{FP8-LM} tries to use FP8 format for gradients by using scaling factors to preserve model convergence; however, \textit{it still easily causes convergence slowdown in training LLMs due to the accuracy loss of gradient accumulation in FP8}.

To this end, in this work, we aim to push activation quantization and gradient quantization a step further and make them practical in LLM training. Specifically, we propose \modelname~with \underline{A}ctivation quantization that can use approximate 4-bit precision and \underline{G}radient quantization with 8-bit for memory-efficient storage and communication-efficient collective, which are compatible with \underline{o}ptimizer state \underline{Q}uantization~\cite{dettmers2022bit} without sacrificing model convergence. To preserve model accuracy and improve system throughput, \modelname~is equipped with our several novel techniques: 1) a layer-aware activation quantization (LAAQ) algorithm (\S\ref{sec:actquant}) that assigns a proper number of bits for activations of different layers according to their layer types and PP stages, which achieves near 4 bits for each element of activation, 2) a precision-preserved quantized gradient storage and communication algorithm named QuanGrad (\S\ref{sec:gradquant}) that uses 8-bit representation (FP8) to store gradients for local accumulation to save memory and for All-Reduce communication to reduce communication time. We conduct extensive experiments using different sizes (from 8B to 34B) of LLMs on a 64-GPU cluster, and the experimental results show that our \modelname{} reduces memory by up to 52\% and achieves a 1.34$\times$ improvement in training speed without sacrificing training loss or accuracy compared to state-of-the-art training systems, including Megatron-LM (w/ or w/o ZeRO), DeepSpeed, and COAT.
\section{Preliminaries}
\subsection{Transformer Layer}\label{subsec:transformer-layer}
Currently, LLMs with transformer architecture~\cite{vaswani2017attention} are the most popular, and a transformer is typically composed of multiple stacked transformer layers. One transformer layer consists of two main sub-components: a self-attention mechanism and a Multi-Layer Perceptron (MLP) typically with two Feed Forward Networks (FFNs). To help easily understand our error analysis in \S\ref{sec:error_ana}, we illustrate their equations.

\textbf{Attention}
The attention layer~\cite{vaswani2017attention} consists of several linear layers that project the input into queries (Q), keys (K), and values (V). The scaled dot-product attention is then computed as:
\begin{equation}\label{equ:attention}
\text{Attention}(Q, K, V)= \text{softmax}\left(\frac{QK^T}{\sqrt{d}}\right)V,
\end{equation}
where $d$ is the dimension of the key vectors. 

\textbf{MLP}
The multi-layer perceptron (MLP) block in Transformer layers consists of two linear transformations with a non-linear activation function in between. 
Typically, the MLP first expands the feature dimension from $M$ to $4M$ (or $8M$ on LLaMa models~\cite{dubey2024llama}) with the first linear layer and then projects it back to the original dimension, which can be represented as:
\begin{equation}\label{equ:mlp}
\text{MLP}(X) = W_2 \times \text{actfunc}(W_1 \times X),
\end{equation}
where $W_1\in \mathbb{R}^{M\times 4M}$, $W_2\in \mathbb{R}^{4M\times M}$ are weight matrices, and actfunc is an activation function like SiLU~\cite{ramachandran2017searching}.

\textbf{SiLU}
The Sigmoid Linear Unit (SiLU) is a smooth, non-monotonic activation function defined as:
\begin{equation}\label{equ:silu}
\text{SiLU}(X) = X \odot \sigma(X),
\end{equation}
where $\odot$ denotes the Hadamard product and $\sigma(X)$ is the sigmoid function. 

\textbf{LayerNorm}
Layer normalization (LayerNorm)~\cite{ba2016layer} is applied to normalize the inputs across the feature dimension for each token independently. RMSNorm~\cite{zhang2019root} is one of the most famous LayerNorm functions:
\begin{equation}\label{equ:rmsnorm}
\text{RMSNorm}(X) = \gamma \frac{X}{\sqrt{\frac{1}{d}\|X\|_2^2 + \epsilon}},  
\end{equation}
where $\|X\|_2^2 = \sum X_i^2$, $\gamma$ is a trainable parameter, $d$ is the number of elements of $X$, and $\epsilon$ is a small constant for numerical stability.

For each hidden layer, its input $X$ must be saved during the forward pass and reused during backpropagation to compute gradients with respect to activations and, for trainable layers, weights. This requirement leads to substantial memory usage. Since different layers perform different types of computations, we observe that compressing $X$ into a low-bit format can introduce varying and potentially large errors in the resulting gradient calculations (details in \S\ref{sec:error_ana}).

\subsection{Paradigms of Parallelism}\label{subsec:parallelism}

\textbf{Data Parallelism (DP)} distributes a mini-batch of samples to multiple workers. During backpropagation, the gradients of each worker in the same DP group are aggregated through an All-Reduce operation so that they can use the identical gradient to update the model parameters.
The All-Reduce operation accumulates distributed gradients (say $X_i$ at $i$ worker) from all workers (say $P$ workers) using a reduction operation (typically sum or mean in training), which can be formally represented
\begin{equation}
    X=\text{AllReduce}(X_1, X_2, ..., X_P)=\sum_{i=1}^{P}X_i.
\end{equation}
The gradients have the same dimensionality as the model weights, which means additional memory is required to store them for communication and updating the model. Compressing the gradient using 8-bit easily causes accuracy loss due to the data overflow of the summation of AllReduce~\cite{FP8-LM}.


\begin{figure}[!t]
	\centering
		\includegraphics[width=0.48\textwidth]{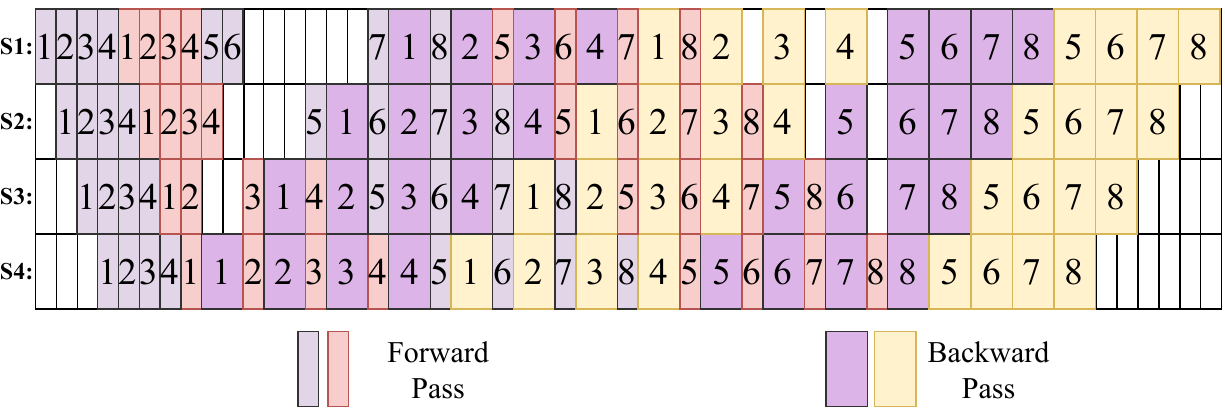}
	\caption{An example of Interleaved 1F1B PP with four stages and each mini-batch divided into eight micro-batches.}
	\label{fig:inter1f1b}
\end{figure}

\textbf{Pipeline Parallelism (PP)}~\cite{huang2019gpipe,narayanan2021efficient} is a commonly used model partitioning strategy for distributed traininge. In PP, the layers of the model are distributed across multiple devices. For models composed of repeated transformer blocks, this typically means assigning an equal number of consecutive transformer layers to each device.
To leverage parallelism within a batch, each batch is further divided into smaller mini-batches. The execution of these mini-batches is then pipelined across devices, overlapping the process of different transformer layers on different devices. To reduce the bubbles, the variant PP known as Interleaved 1F1B~\cite{narayanan2021efficient}, as illustrated in Fig.~\ref{fig:inter1f1b}, has been widely used in Megatron-LM. In this scheme, after a mini-batch passes through the entire device sequence (from the first to the last device), it is sent back to the first device and traverses the device sequence again. This requires partitioning the model into more granular segments and placing these segments evenly across devices according to the order in which the mini-batch will traverse them.  

Additionally, each forward pass in Fig.~\ref{fig:inter1f1b} stores a portion of activations in GPU memory for the subsequent backward pass, while each backward pass releases part of these activations after computation. When PP is employed, the amount of concurrently stored activations differs across devices, which will be discussed in Section~\ref{sec:dbc}.
\begin{figure}[!t]
	\centering
		\includegraphics[width=0.48\textwidth]{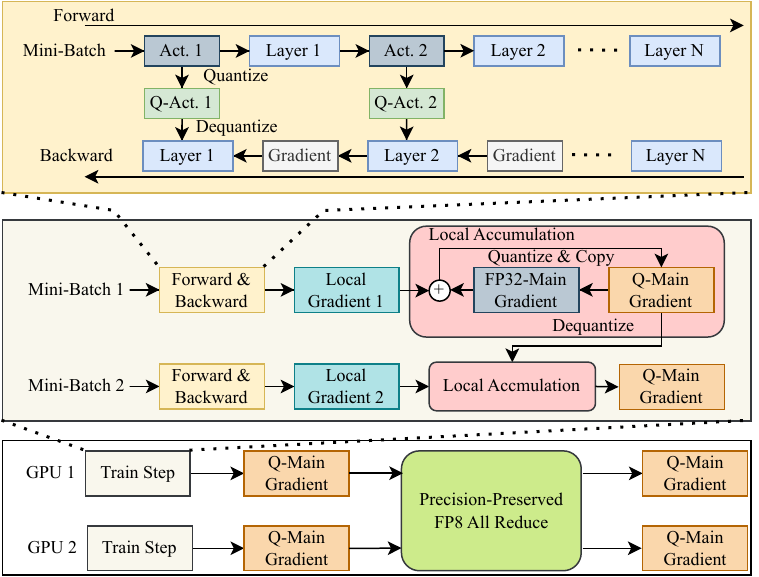}
	\caption{The integration of activation and gradient quantization with Megatron-LM. }
	\label{fig:quant_process}
\end{figure}
\begin{figure}[!t]
	\centering
		\includegraphics[width=0.4\textwidth]{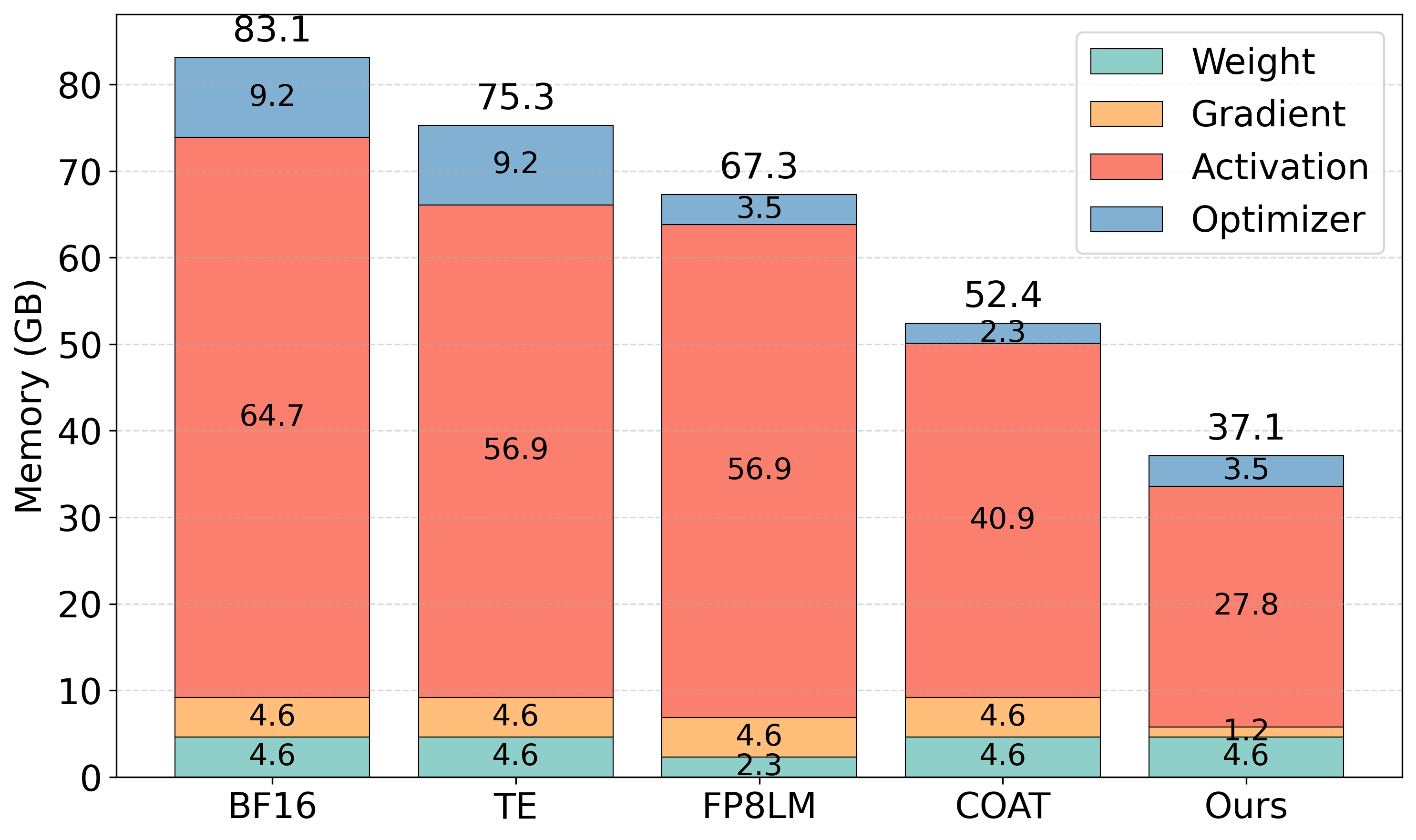}
	\caption{Training memory consumption on an OLMo-1B model.}
	\label{fig:memory}
\end{figure}

\section{\modelname{}: System Overview}
To significantly reduce the GPU memory footprint of LLM training, we design our \modelname{} to compress activations to nearly 4 bits and gradients to 8 bits, which is also compatible with the 8-bit Adam optimizer~\cite{dettmers2022bit} atop Megatron-LM. As shown in Fig.~\ref{fig:quant_process}, \modelname{} introduces two new components: nearly 4-bit activation quantization and 8-bit gradient quantization. 

First, for activation quantization, the forward pass first generates full-precision activations, which are then quantized and stored in approximately 4-bit precision. During the backward pass, the quantized activations are dequantized to BF16/FP16 before gradient computation. In principle, 4-bit activations require only one quarter of the memory of FP16/BF16. However, naively quantizing the activations of all layers to 4 bits leads to substantial accuracy degradation. To gain the memory advantages of 4-bit while maintaining model performance, we introduce layer-aware activation quantization (\S\ref{sec:actquant}).

Second, for gradient quantization, at each GPU, the local gradient is first computed via forward and backward passes per mini‑batch, then accumulated with the main gradient through local gradient accumulation. This process involves dequantizing the quantized main gradient (Q‑Main Gradient), adding it to the local gradient in high precision, and quantizing the sum back to 8‑bit format before copying it to the Q‑Main Gradient. After local accumulation, GPUs perform a precision‑preserved FP8 All‑Reduce across GPUs, as described in \S\ref{sec:gradquant}.

As shown in Fig.~\ref{fig:memory} with an OLMo-1B model~\cite{OLMo}, we compare the memory footprint across different components for the BF16 baseline, Transformer Engine (TE)~\cite{TransformerEngine2024}, FP8-LM~\cite{peng2023fp8}, COAT~\cite{xi2024coat}, and our method. AGoQ achieves further compression on both activations (\S\ref{sec:actquant}) and gradients (\S\ref{sec:gradquant}): compared to COAT, we reduce activation memory by an additional 30\% and gradient memory by 75\%.


\section{Layer-Aware Activation Quantization}\label{sec:actquant}
To minimize accuracy degradation while maximizing overall memory savings, we first identify which layers’ activations are appropriate for 4-bit compression through a theoretical analysis, since different layer types (e.g., Attention, FFN, LayerNorm) exhibit distinct computation patterns as introduced in \S\ref{subsec:transformer-layer}. Second, the PP training paradigm leads to uneven memory usage across different PP stages, which can be leveraged to design a dynamic quantization compensation strategy that takes advantage of underutilized memory resources.

\begin{table*}[!htb]
\centering
\caption{Activation memory of different operations. U is a unit to measure memory usage, where 1U = Batch Size × Sequence Length × Hidden Size × 2 bytes (for BF16). Act Func refers to SiLU \& Multiply.}
\label{tab:memory_comparison}
\begin{tabular}{lcccccccc}
\toprule
& \textbf{QKV} & \textbf{Attention} & \textbf{Linear} & \textbf{RMSNorm} & \textbf{FFN1} & \textbf{Act Func} & \textbf{FFN2} & \textbf{Total} \\
\midrule
Megatron-LM (w/ BF16) & 1U & 5U & 1U & 4U & 1U & 12U & 4U & 28U \\
COAT & 1U & 5U & 1U  & 1U & 0.5U & 6U & 2U & 16.5U \\
\modelname{} & 0 & 5U & 0.25U & 0.5U & 0 & 2U & 0 & 7.75U \\
\bottomrule
\end{tabular}
\end{table*}
\subsection{Error Analysis of Activation Quantization}\label{sec:error_ana}
To minimize accuracy loss, we perform a numerical analysis to determine which activations should be quantized for different types of layers. 


We first categorize different modules into two types based on whether they need to save additional activations beyond the input during computation. The matrix multiplication (GEMM) module in MLP only needs to save input activations. The modules that require saving additional activations include RMSNorm, SiLU \& Multiply and attention modules. 
To illustrate the difference between the two types, we take RMSNorm (Eq.~\ref{equ:rmsnorm}) as an example.
Let
\begin{equation}
r = \sqrt{\frac{1}{d}\|X\|_2^2 + \epsilon},
\end{equation}
then it can be written as \( Y =\text{RMSNorm}(X) = \gamma X / r \). The gradient matrix is expressed as:
\begin{equation}
J = \frac{\operatorname{diag}(\gamma)}{r} - \frac{1}{d} \frac{\operatorname{diag}(\gamma) X X^T}{r^3}.
\label{equ:rmsnorm_g}
\end{equation}
Therefore, to compute the gradient, we need to store both $X$ and $r$. Here, $r$ represents the additional activations that should also be cached. When using recomputation techniques, we do not store $r$; instead, during the backward pass, $r$ is recomputed from $X$ before the gradient calculation.

For modules requiring additional activations, we primarily analyze gradient errors under two scenarios:

\textbf{Case 1 (Recompute intermediate values)}: Only the quantized input activations are stored, and the originally required additional activations are recomputed during gradient calculation using the quantized input activations.

\textbf{Case 2 (Cache intermediate values)}: Both the quantized input activations and the quantized additional activations are stored.

When analyzing GEMM computations adjacent to operations like RMSNorm or SiLU, we specifically compare two gradient computation strategies: one where only the quantized inputs to the preceding operation (e.g., RMSNorm/SiLU) are stored and the GEMM inputs are recomputed from these quantized values during backpropagation (also called Case 1), versus an alternative approach where the GEMM inputs themselves are directly stored in quantized form to avoid recomputation (Case 2). 

In the following derivations, we make use of three standard norm inequalities. 
For vectors $X,Y \in \mathbb{R}^d$, the $\ell_2$-norm of their element-wise product satisfies
\begin{equation}
\|X \odot Y\|_{2} \leq \|X\|_{2} \, \|Y\|_{\infty}.
\label{eq:elementwise-ineq}
\end{equation}
For matrices $A \in \mathbb{R}^{m \times k}$ and $B \in \mathbb{R}^{k \times n}$, the spectral norm is sub-multiplicative:
\begin{align}
\|AB\|_{2} \leq \|A\|_{2} \, \|B\|_{2} \\
\|AB\|_{2} \leq \|A\|_{2} \, \|B\|_{\infty}
\label{eq:product_imba}
\end{align}

\subsubsection{RMSNorm}
RMSNorm and its gradient are defined as Eq.~\ref{equ:rmsnorm} to Eq.~\ref{equ:rmsnorm_g}.

\textbf{Case 1 (Recompute intermediate values):} Only the quantized input $x$ is stored. We assume the error introduced by quantization can be modeled as a multiplicative perturbation:
\begin{equation}
X' = X \odot (1 + \delta X), \quad r' = r(X'), \quad J' = J(X').
\end{equation}
Let \( \Delta J = J' - J \). Performing first-order expansion:
\begin{equation}
r' = r + \Delta r, \quad \Delta r \approx \frac{1}{2r} \frac{2}{d} \sum X_i^2 \delta X_i.
\label{eq:deltar}
\end{equation}

We obtain:
\begin{equation}
\Delta\left( \frac{1}{r} \right) \approx -\frac{\Delta r}{r^2}, \quad \Delta\left( \frac{1}{r^3} \right) \approx -3\frac{\Delta r}{r^4}.
\end{equation}
Denoting $D_\gamma = \text{diag}(\gamma)$ and substituting into \(\Delta J\), we have
\begin{equation}
\small
    \begin{aligned}
        \Delta J &\approx D_\gamma \Bigg[ \Delta\!\left(\frac{1}{r}\right) I \nonumber  - \frac{1}{d} \bigg( \Delta\!\left(\frac{1}{r^3}\right) X X^T + \frac{1}{r^3} \Delta(X X^T) \bigg) \Bigg] \nonumber \\
&= -\frac{\Delta r}{r^2} D_\gamma + \frac{3\Delta r}{d r^4} D_\gamma X X^T \nonumber \\
&\quad - \frac{1}{d r^3} D_\gamma \left[ X (\delta X \odot X)^T + (\delta X \odot X) X^T \right]
    \end{aligned}
\end{equation}
Using Eq.~\ref{eq:elementwise-ineq} to Eq.~\ref{eq:product_imba}, we have
\begin{equation}
\small
    \begin{aligned}
        \|\Delta J\|_2 &\lesssim \|\gamma\|_\infty \Bigg( \frac{|\Delta r|}{r^2} + \frac{3|\Delta r|}{d r^4} \|X\|_2^2 + \frac{2}{d r^3} \|X\|_2^2 \|\delta X\|_\infty \Bigg),
    \end{aligned}
\end{equation}
where \( ||\gamma||_\infty = \max |\gamma_i| \). Further substituting Eq.~\ref{eq:deltar}, then we have
\begin{equation}\label{equ:rmsnorm-case1-error}
\|\Delta J\|_2 \lesssim  3  ||\gamma||_\infty \|\delta X\|_\infty \left( \frac{\|X\|_2^2}{d r^3} + \frac{\|X\|_2^4}{d^2 r^5} \right) .
\end{equation}
Typically \( \epsilon \) is constant and \( r \approx \sqrt{\|X\|^2_2 / d} \), so the leading order is \( \mathcal{O}(||\gamma||_\infty \|\delta X\|_\infty /r) \).

\textbf{Case 2 (Cache intermediate values): } Both the quantized input $X$ and the quantized additional activation $r$ are stored. In this case, the error is expressed as:
 \(X' = X \odot (1+\delta X)\) and \(r' = r(1+\delta r)\) be quantized, with \(|\delta r|,\|\delta X\|_\infty \le \varepsilon_q\).  
The perturbed Jacobian is  
\begin{equation}
    J' = \frac{D_\gamma}{r'} - \frac{1}{d}\frac{D_\gamma X'(X')^T}{(r')^3}.
\end{equation}
Expanding to first order:
\begin{equation}
\begin{aligned}
    \frac{1}{r'} \approx \frac{1}{r}(1-\delta r),\quad
\frac{1}{(r')^3} \approx \frac{1}{r^3}(1-3\delta r),\quad \\
X'(X')^T \approx XX^T + X(X\odot\delta X)^T + (X\odot\delta X)X^T.
\end{aligned}
\end{equation}
Thus  
\begin{equation}
\small
\begin{aligned}
J' &\approx \frac{D_\gamma}{r} - \frac{\delta r}{r}D_\gamma + \frac{3\delta r}{d r^3}D_\gamma XX^T \\
&\quad - \frac{1}{d r^3}D_\gamma\Big[XX^T + X(X\odot\delta X)^T + (X\odot\delta X)X^T\Big].
\end{aligned}
\end{equation}
Subtracting the exact \(J = \frac{D_\gamma}{r} - \frac{1}{d r^3}D_\gamma XX^T\) yields the first‑order perturbation:
\begin{equation}
\small
\begin{aligned}
    \Delta J &\approx 
-\frac{\delta r}{r}D_\gamma
- \frac{1}{d r^3}D_\gamma\big[X(X\odot\delta X)^T + (X\odot\delta X)X^T\big] \\
&\quad + \frac{3\delta r}{d r^3}D_\gamma XX^T .
\end{aligned}
\end{equation}
Taking 2‑norm bounds by Eq.~\ref{eq:elementwise-ineq} to Eq.~\ref{eq:product_imba} gives
\begin{equation}
    \|\Delta J\|_2 \;\lesssim\; \frac{6\|\gamma\|_\infty\,\varepsilon_q}{r}.
    \label{equ:rmsnorm-case2-error}
\end{equation}

According to Eq.~\ref{equ:rmsnorm-case1-error} and Eq.~\ref{equ:rmsnorm-case2-error}, we conclude that storing only the quantized input activations and recomputing intermediate values during gradient calculation yields the same asymptotic error order as caching intermediate values, with only constant-factor differences — and notably, the constant factor for the recomputation approach appears tighter. 
\textbf{Given that recompute also reduces memory footprint, for RMSNorm, we store the input activations as 4-bit and recompute their intermediate values for gradient computation.}

\subsubsection{Other Operations}
The detailed analyses of other operations like SiLU \& Multiply, RMSNorm+GEMM, and Attention have similar derivations to RMSNorm, so they are provided in Appendix~\ref{ap:error_analy} and we conclude the following results.

We apply the Case 1 recomputation strategy to $Q$, $K$, $V$, intermediate activations of RMSNorm and SiLU \& Multiply, input activations of two FFNs in MLP of each transformer layer, eliminating storage for those values. Based on the analysis in Appendix~\ref{ap:error_analy}, the gradient error of quantizing activations of attention modules is significantly larger than that of other modules, so we do not quantize the activations of attention modules. Other activations are quantized to 4 bits via block‑wise FP4 quantization~\cite{dettmers2022bit,li2023memory} with the blocksize of 128. As shown in Table~\ref{tab:memory_comparison}, our approach reduces RMSNorm memory from 4U\footnote{Note that in Megatron-LM with BF16 training, RMSNorm still uses FP32 for better convergence.} to 0.5U (4‑bit), and SiLU \& Multiply activations from 12U to 2U. Attention remains at 5U as reported in \cite{xi2024coat}. Overall, activation memory drops from 28U in Megatron‑LM to 7.75U in our method—an approximate three‑fold reduction.

\subsection{Dynamic Bit-width Compensation}\label{sec:dbc}
Interleaved 1F1B PP leads to imbalanced memory footprints across devices. As illustrated in Fig.~\ref{fig:inter1f1b} (four PP stages with 8 mini‑batches), different devices store varying numbers of activation batches—e.g., Device 1 holds 11 mini‑batch activations at peak, while Devices 2, 3, and 4 store only 9, 7, and 5, respectively. This results in significant under‑utilization of GPU memory, with Device 1 occupying 2.2$\times$ more activation memory than Device 4.

To exploit the under-utilized memory, we propose Dynamic Bit‑width Compensation for Activation with Pipeline Parallelism (DBCA‑PP). Devices storing fewer activation batches are assigned higher quantization bit‑widths, thereby compensating for quantization‑induced precision loss without increasing peak memory usage. This strategy makes full use of the otherwise wasted memory across the pipeline, while keeping nearly 4-bit activation storage.

Formally, in Interleaved 1F1B PP, the number of activation mini-batches $N_i$ stored at stage $i$ (with totally $n$ stages) can be expressed as:
\begin{equation}
    N_i = n+ 2 \cdot i-1,\quad 1 \le i \le n.
\end{equation}
Based on the memory availability per stage, the quantization bit-width $B_i$ (with a minimum of 4) for activations at stage $i$ can be set as inversely proportional to the number of activation mini-batches $N_i$:
\begin{equation}
    B_i = \frac{4\cdot N_{1}}{N_i},\quad 1 \le i \le n.
    \label{dynamic}
\end{equation}
It is worth mentioning that the bit-width configuration generated for a lower number of stages can also be directly applied to a setup with a higher number of stages. In other words, directly reusing the bit assignment scheme from a lower-stage configuration in a higher-stage setup with DBCA-PP would not increase the peak GPU memory usage of the higher-stage setup.

\begin{figure}[!t]
	\centering
 \begin{subfigure}[b]{0.35\textwidth}
		\centering
		\includegraphics[width=\textwidth]{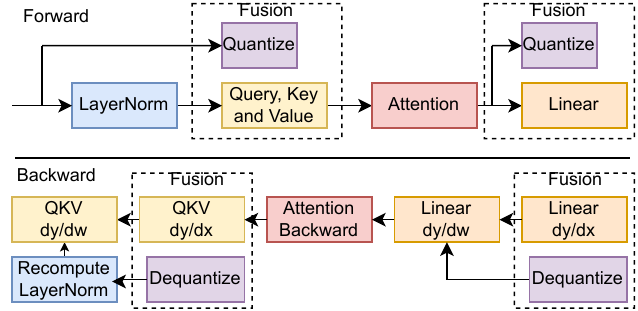}
		\caption{Attention}
		\label{fig:att}
	\end{subfigure}
    
	 \begin{subfigure}[b]{0.35\textwidth}
		\centering
		\includegraphics[width=\textwidth]{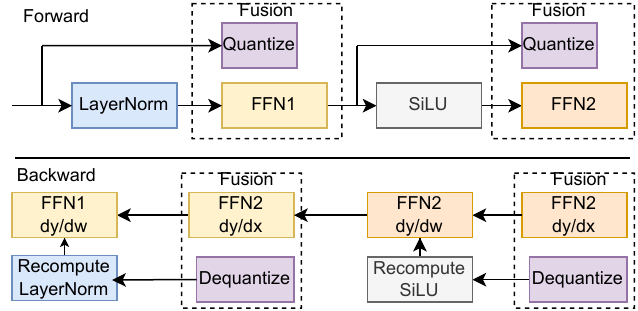}
		\caption{MLP}
		\label{fig:mlp}
	\end{subfigure}
    
	\caption{The forward and backward passes of attention and MLP for kernel fusion of quantization/dequantization and GEMM.}
	\label{fig:fuse}
\end{figure}

\subsection{Kernel Fusion of Quantization/Dequantization and GEMM}\label{subsec:kernelfusion}
The activation quantization and dequantization require extra computation overheads. To address this problem, we fuse these operations along with nearby GEMM computations into a single GPU kernel. This is motivated by the fact that quantization and dequantization are mainly element-wise operations, thus only utilize CUDA cores, whereas GEMM leverages Tensor Cores on modern GPUs. 

To achieve this goal, we carefully schedule the execution of activation quantization, dequantization, and GEMM operations during LLM training as shown in Fig.~\ref{fig:fuse}. During the forward pass, we fuse the quantization process with its subsequent GEMM operation. During the backward pass, we fuse dequantization with the GEMM operation that is responsible for computing activation gradients. This approach can almost eliminate the computation overheads of activation quantization, thus improving execution efficiency.

\begin{figure}[!t]
	\centering
		\includegraphics[width=0.35\textwidth]{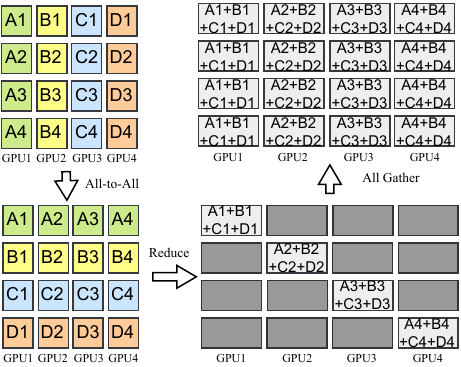}
	\caption{The illustration of our process to perform All-Reduce by combining All-to-All with All-Gather.}
	\label{fig:a2a_ag}
\end{figure}

\begin{figure}[!t]
	\centering
		\includegraphics[width=0.48\textwidth]{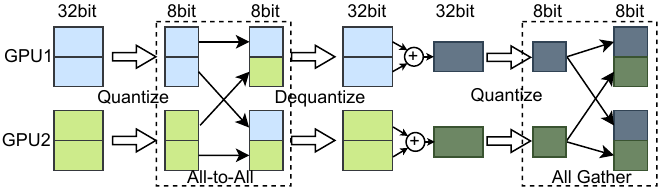}
	\caption{The illustration of our process to combine gradients from different GPUs.}
	\label{fig:grad_a2a}
\end{figure}

\section{Precision-Preserved Gradient Quantization}\label{sec:gradquant}

To minimize both the memory usage associated with storing gradients and the communication overhead during gradient All-Reduce, we introduce an 8-bit block-wise~\cite{dettmers2022bit} gradient quantization technique. This method maintains precision throughout the All-Reduce operation and effectively mitigates two distinct overflow issues found within the accumulation process.

First, in LLM training, a global batch is typically divided into multiple mini-batches whose gradients are locally accumulated before communication with other DP workers. When we store the main gradients with FP8, directly accumulating gradients from different mini-batches would easily cause overflow. Thus, in local gradient accumulation, we dequantize the FP8 main gradient to FP16/BF16 for high-precision addition of different mini-batch gradients, and the final results are then quantized to FP8 as shown in Fig.~\ref{fig:quant_process}. 

Second, the gradients should be aggregated among DP workers via an All-Reduce operation, which can be divided into Reduce-Scatter and All-Gather. However, Reduce-Scatter requires performing addition during communication, which could easily cause an overflow in FP8. Thus, we split the All-Reduce operation into an All-to-All operation with a local reduce followed by an All-Gather operation as shown in Fig.~\ref{fig:a2a_ag}.
As shown in Fig.~\ref{fig:grad_a2a}, the FP8 gradients are communicated via an All-to-All communication to send the compressed data to all devices. Each device then dequantizes the data received from different devices to FP32, performs local reduce operations, and then quantizes the result again for its following All-Gather operation. After that, we perform All-Gather on the summed data to complete the All-Reduce operation. 


\section{Evaluation}\label{sec:evaluation}



\subsection{Experimental Settings}
\textbf{Testbeds.}
Experiments are mainly carried out on a 64-GPU cluster connected with 200Gb/s InfiniBand comprising 8 nodes, each of which is equipped with eight Nvidia A6000 GPUs. In the comparative experiments with COAT~\cite{xi2024coat}, we used two nodes, each equipped with 8 NVIDIA Pro 6000 GPUs, to support the FP8 format required by COAT. The software environments are Ubuntu-20.04, CUDA-12.1, PyTorch-2.1.2, and NCCL-2.18.5. Our system also supports Huawei Ascend 910 NPUs (more results can be found in Appendix~\ref{subsec:ablation}).

\textbf{Baselines.} We implement our \modelname{} atop Megatron-LM. We compare our \modelname{} with three representative baselines Megatron-LM~\cite{narayanan2021efficient} (w/o and w/ ZeRO~\cite{rajbhandari2020zero}), DeepSpeed~\cite{rasley2020deepspeed}, and COAT~\cite{xi2024coat}.

\textbf{Models.} We primarily conduct pre-training experiments to verify convergence on LLaMA2-7B~\cite{touvron2023llama} due to extremely high training costs, and perform training time comparison experiments on larger models including LLaMA3-8B~\cite{dubey2024llama}, LLaMA2-13B~\cite{touvron2023llama}, and CodeLLaMA-34B~\cite{roziere2023code}. When comparing with COAT, we employed the OLMo-1B model~\cite{OLMo} provided by the original COAT paper~\cite{xi2024coat}.

\begin{table}[h]
\centering
\small
\caption{Performance Comparison of \modelname{}, Megatron-LM and ZeRO-1 on LLaMA2-13B. R means the number of recomputed transformer layers. The unit of time is milliseconds (ms).}
\label{tab:performance_comparison}
\begin{tabular}{@{}ccccccc@{}}
\toprule

Sequence & \multicolumn{2}{c}{Megatron-LM} & \multicolumn{2}{c}{ZeRO-1} & \multicolumn{2}{c}{\modelname{}} \\
\cmidrule(lr){2-3}\cmidrule(lr){4-5}\cmidrule(lr){6-7}
Length & R & Time & R & Time & R & Time \\
\midrule
32K & 3 & 37635 & 2 & 37038 & 0 & 36568 \\
40K & 4 & 51922 & 4 & 51418 & 0 & 45590 \\
48K & 6 & 67932 & 6 & 67928 & 0 & 57047 \\
56K & 8 & 88200 & 8 & 86601 & 0 & 69544 \\
64K & 8 & 104444 & 8 & 103706 & 0 & 82519 \\
72K & 10 & 128085 & 10 & 128152 & 0 & 97615 \\
80K & 10 & 149667 & 10 & 149288 & 0 & 111422 \\
\bottomrule
\end{tabular}
\end{table}

\subsection{End-to-end Training Time Comparison}
\textbf{LLaMA2-13B}.
To assess the effectiveness of our method, we compare \modelname{} with Megatron-LM without ZeRO-1 and with ZeRO-1. We benchmark training speed on LLaMA2-13B with sequence lengths from 32K to 80K. Under limited GPU memory, we adopt selective activation recomputation: instead of caching all intermediate activations after the forward pass, we recompute them via an extra forward pass during backpropagation. The number of transformer layers that use recomputation is adaptively tuned based on real-time memory usage. Table~\ref{tab:performance_comparison} reports results on 64 GPUs with mini-batch size 1, global batch size 16, PP=4 and TP=8. Overall, \modelname{} achieves an average speedup of 1.22$\times$ over Megatron-LM and 1.21$\times$ over its ZeRO-1 variant (we focus on ZeRO-1 here because ZeRO-2 and ZeRO-3 introduce additional communication overhead, typically reducing throughput, as shown in \S\ref{subsec:ablation} in Appendix). The speedup grows with sequence length, confirming the benefit of our approach; for instance, at 80K tokens, \modelname{} is about $1.34\times$ faster than both Megatron-LM and ZeRO-1.

\begin{figure}[!t]
	\centering
 \begin{subfigure}[b]{0.22\textwidth}
		\centering
		\includegraphics[width=\textwidth]{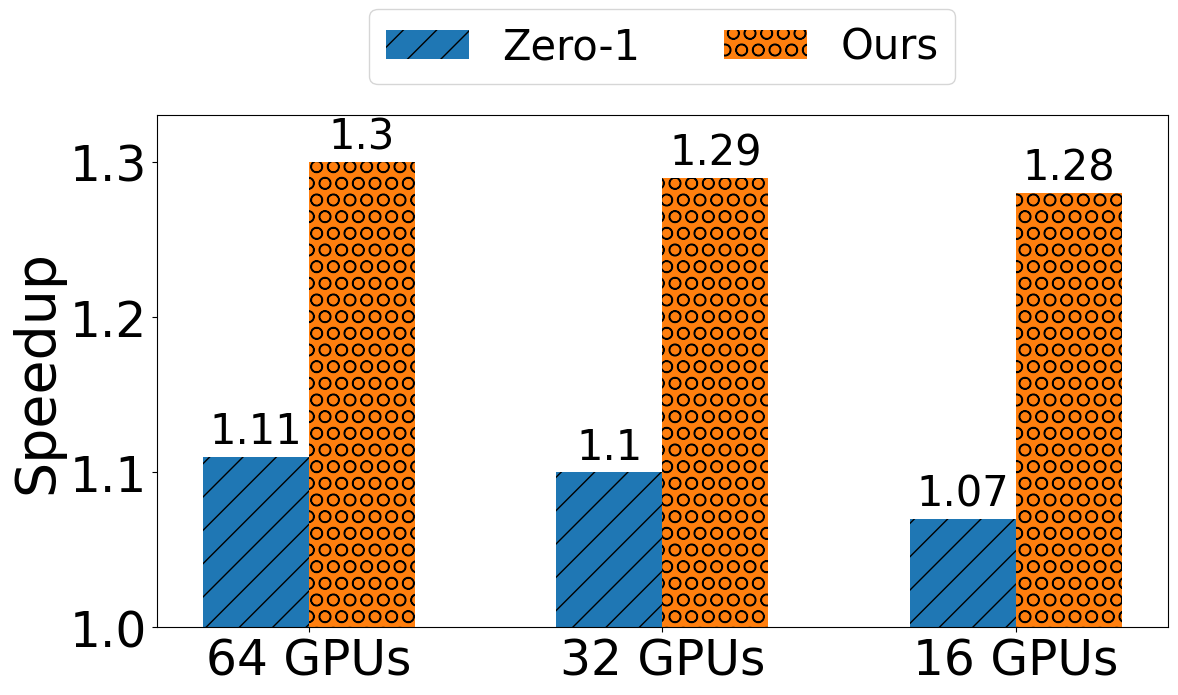}
		\caption{Varied number of GPUs}
		\label{fig:gpus}
	\end{subfigure}
     \begin{subfigure}[b]{0.22\textwidth}
		\centering
		\includegraphics[width=\textwidth]{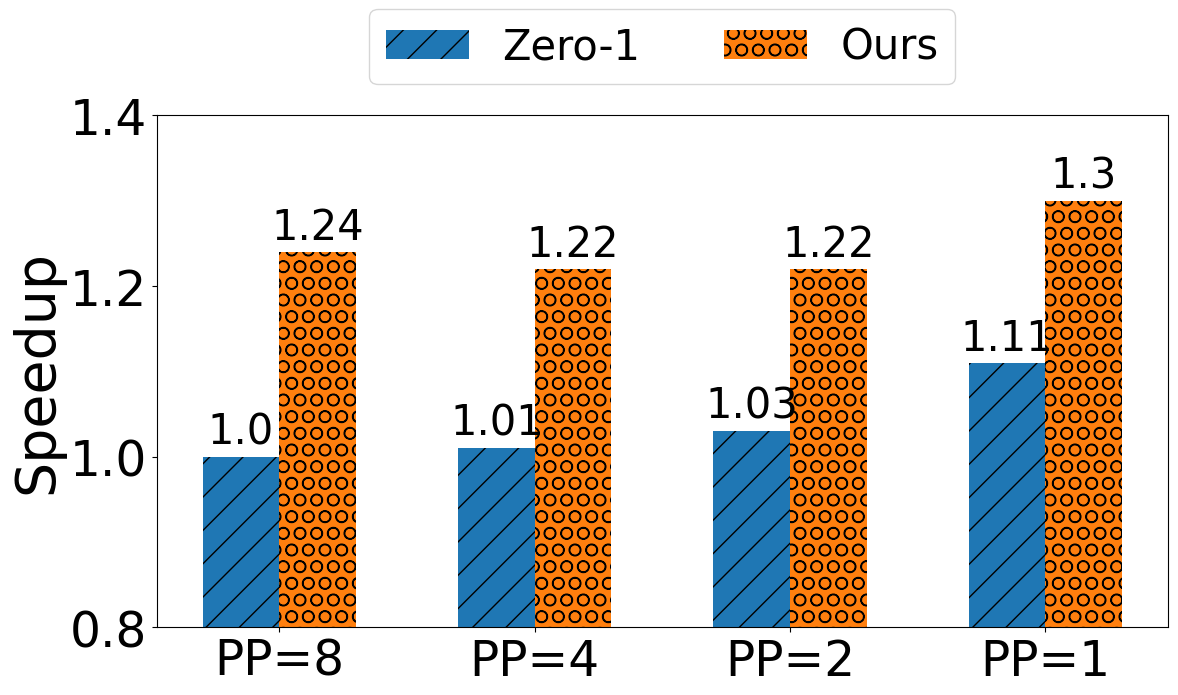}
		\caption{Varied degree of PP}
		\label{fig:pp}
	\end{subfigure}
	\caption{Speedups of our \modelname{} and ZeRO-1 over Megatron-LM on varied configurations.}
	\label{fig:spup}
\end{figure}

\textbf{Performance under Different Configurations}.
To further evaluate our method, we broaden the experimental settings. We vary the sequence length from 16K to 32K or from 32K to 80K, set pipeline parallelism (PP) from 1 up to the number of nodes, tensor parallelism (TP) in {4, 8}, and the number of GPUs in {8, 16, 32, 64}. We test three models: LLaMA3-8B, LLaMA2-13B, and CodeLLaMA-34B. The detailed configurations are summarized in Appendix~\ref{subsec:ablation}, where we run 124 experiments each for Megatron-LM, ZeRO-1, and \modelname{}, for a total of 372 experiments. Of these, 69 runs failed due to out-of-memory (OOM) errors (\modelname{}, Megatron-LM, and ZeRO-1 have 16, 33, and 20 OOM cases, respectively), and 303 runs completed successfully. Overall, the results show that \modelname{} delivers an average speedup of 1.23$\times$ over Megatron-LM and 1.19$\times$ over ZeRO-1. We next examine the impact of GPU count, sequence length, model size, and PP separately. Resutls are shown in Fig.~\ref{fig:spup}, which indicates that \modelname{} consistently achieves substantial improvements over Zero-1 and Megatron-LM across different GPU counts and degree of PP.

\begin{table}[t]
\centering
\small
\caption{Time and memory at different sequence lengths.}
\label{tab:seq_len_comparison}
\begin{tabular}{lccc}
\toprule
Seq. Len. & Method & Time (ms) & Memory (MB) \\
\midrule
\multirow{2}{*}{24k} & COAT & 6291 & 94100 \\
                      & AGoQ & 6161 & 66852 \\
\midrule
\multirow{2}{*}{32k} & COAT & 8861 & 95664 \\
                      & AGoQ & 8076 & 86012 \\
\bottomrule
\end{tabular}
\end{table}
\textbf{Comparison with COAT}.
We further compare \modelname{} with COAT, using two nodes with 8 Pro6000 GPUs to support COAT's FP8 format. With a global batch size of 64 and sequence lengths of 24K and 32K, we evaluate the OLMo-1B model. For 32K sequences, COAT encounters OOM errors, requiring recomputation for half of the transformer layers. Results in Table~\ref{tab:seq_len_comparison} show that at 24K, our \modelname{} reduces memory by 31\% over COAT while matching training speed; at 32K, with recomputation enabled for COAT, \modelname{} achieves a 1.1$\times$ end-to-end speedup. Due to our hardware limit, we only conduct the experiments on 16 Blackwell GPUs. It is expected to achieve higher speedups over COAT on larger clusters since \modelname{} allows 8-bit communication to significantly reduce DP communication overhead.

\begin{figure}[!t]
	\centering
		\begin{subfigure}[b]{0.22\textwidth}
		\centering
		\includegraphics[width=\textwidth]{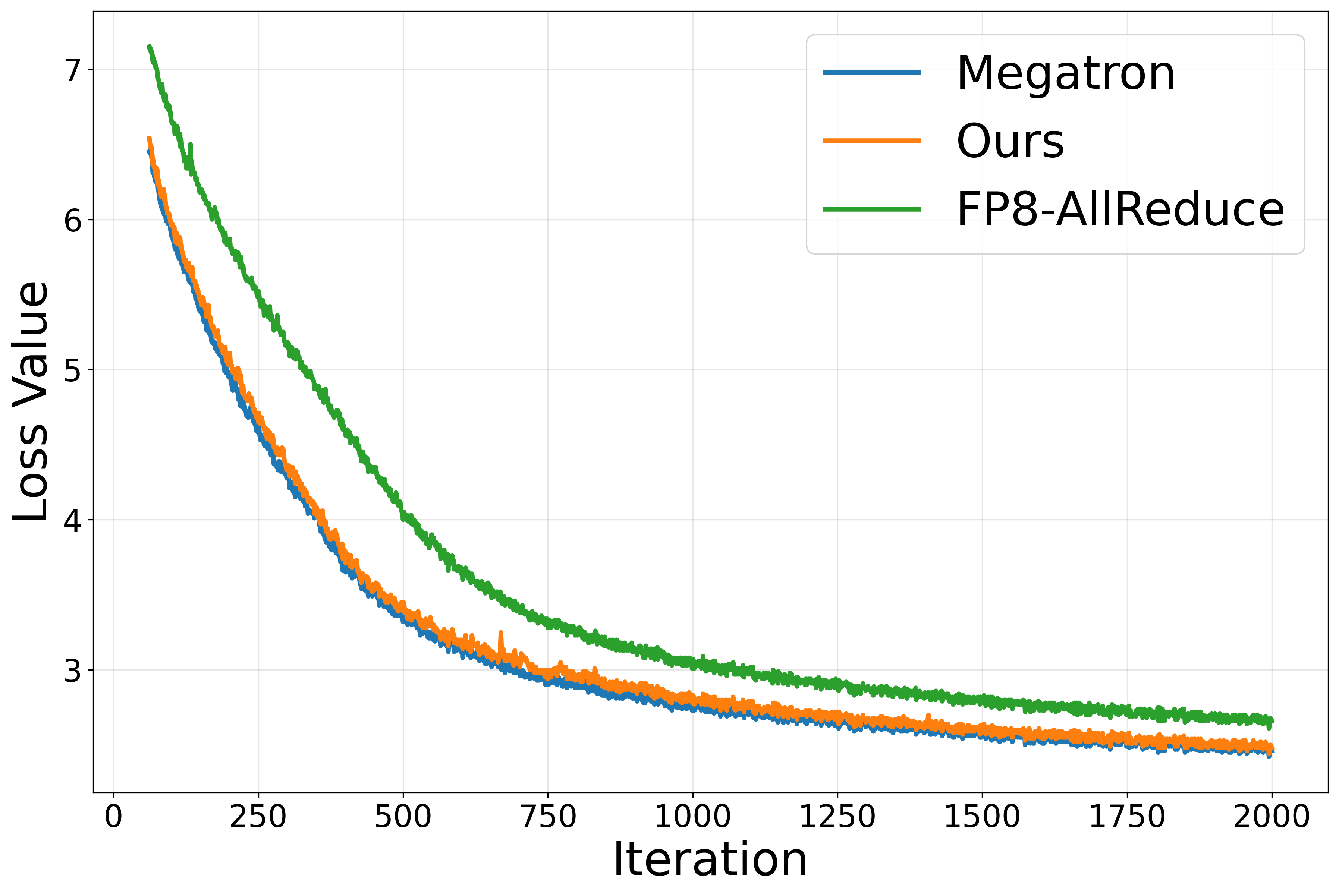}
		\caption{Llama2-7B}
		\label{fig:cur_llama2}
	\end{subfigure}
     \begin{subfigure}[b]{0.22\textwidth}
		\centering
		\includegraphics[width=\textwidth]{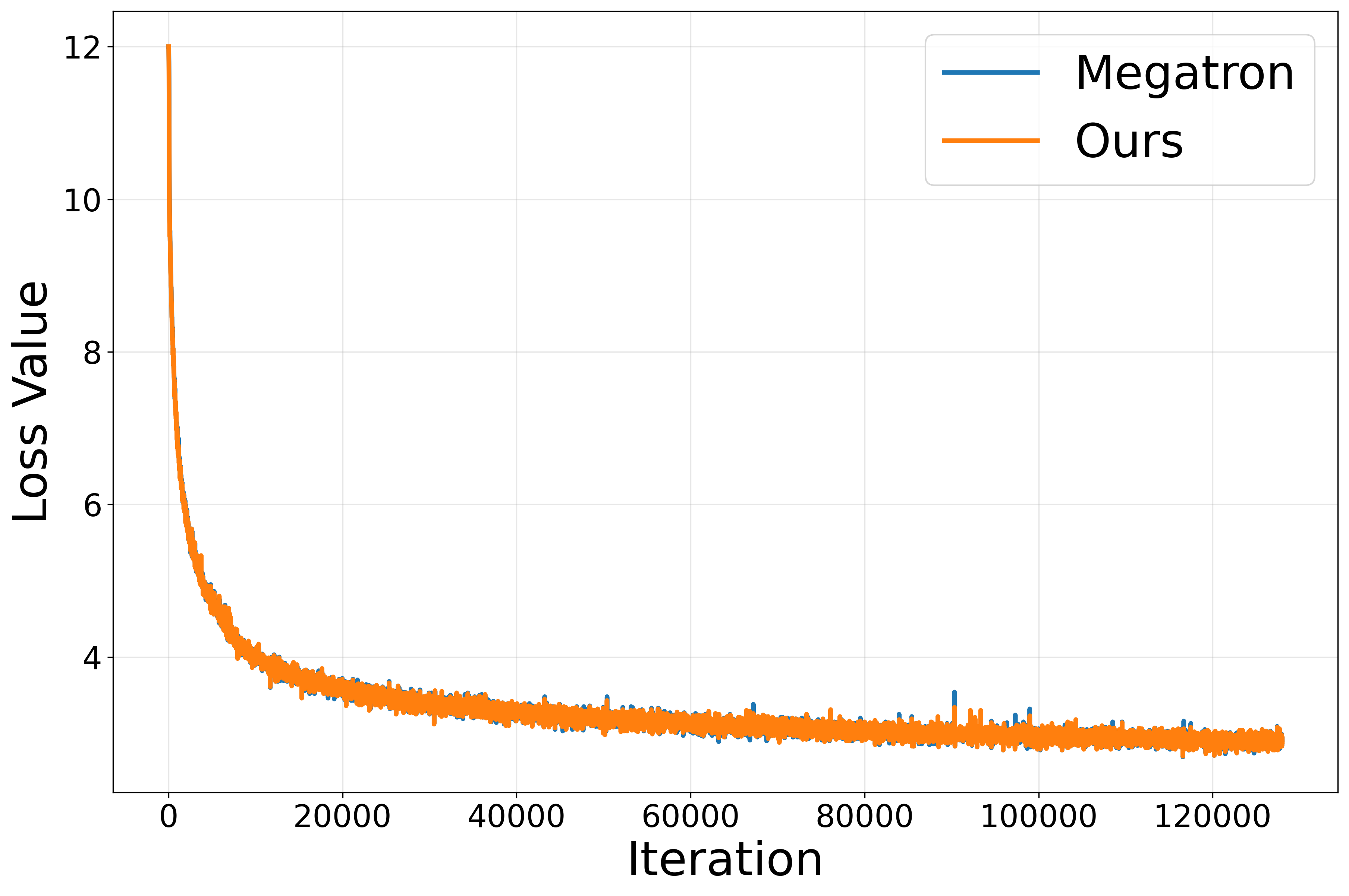}
		\caption{Llama3-8B}
		\label{fig:cur_llama3}
	\end{subfigure}
	\caption{Training loss of Megatron (w/ BF16), FP8-AllReduce and ours (A+O+G) on LLaMA2-7B and LLaMA3-8B.}
	\label{fig:loss}
\end{figure}
\subsection{Convergence Loss}
We evaluate the convergence of \modelname{} by pretraining LLaMA2-7B and LLaMA3-8B on 2B tokens from OpenWebText~\cite{peterson2019open}. We evaluate the robustness of our method by testing it with different global batch sizes on two models: 512 in LLaMA2‑7B and 4 in LLaMA3-8B. Using an interleaved 1F1B schedule with 4 pipeline stages, we apply DBCA-PP with activation bit-widths of 4, 5, 6, and 8 per stage according to Eq.~\ref{dynamic}, matching the peak memory footprint of uniform 4-bit compression. During the training of LLaMA2‑7B, we additionally tested the training curve of FP8‑AllReduce (Microsoft's FP8 AllReduce method~\cite{FP8-LM}). As shown in Fig.~\ref{fig:loss}, our approach closely track the baseline loss, while FP8-AllReduce shows significantly higher loss.

\begin{figure}[!t]
	\centering
		\includegraphics[width=0.3\textwidth]{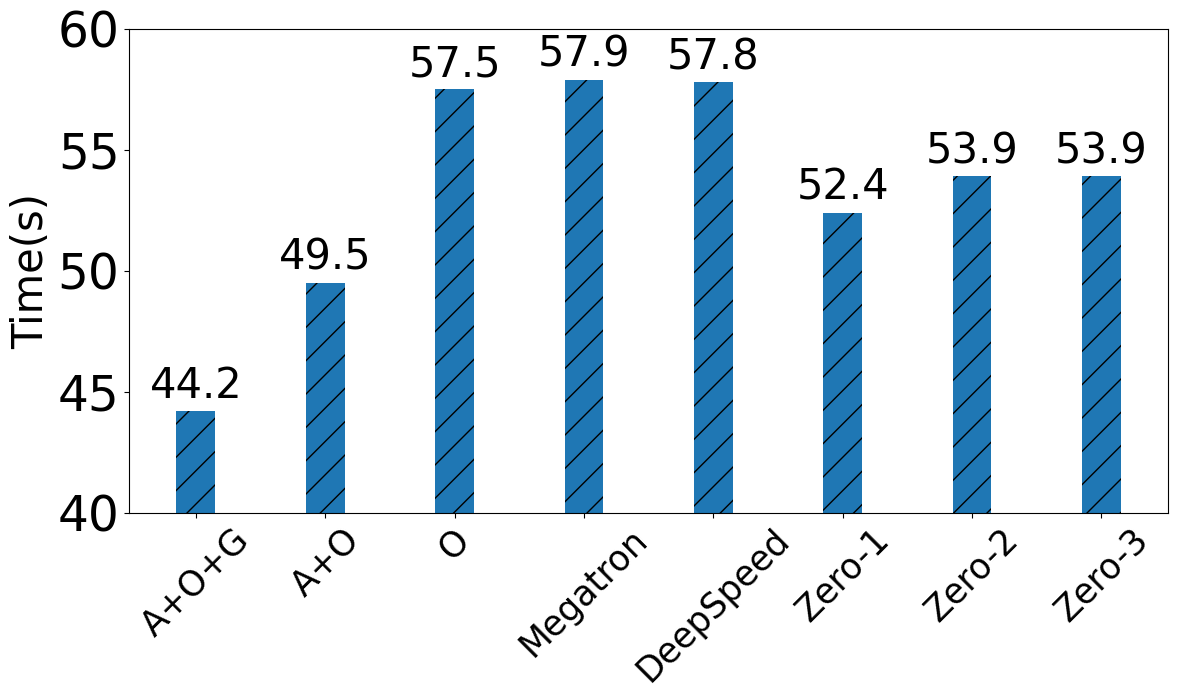}
	\caption{Iteration time (in seconds) comparison on LLaMA2-13B.}
	\label{fig:zero}
\end{figure}
\textbf{Comparison of Different Optimizations}.
We also validated the individual contributions of activation quantization, gradient quantization, and optimizer quantization modules, while simultaneously examining the differences between DeepSpeed, ZeRO-1, ZeRO-2, and ZeRO-3. Under the configuration of PP=1 and sequence length=48K, we conducted experiments on LLaMA2-13B with three different settings: ``A+O+G'' (i.e., AGoQ), ``A+O'' (only applying activation and optimizer quantization) and ``O'' (only applying 8-bit optimizer quantization) shown in Fig.~\ref{fig:zero}, demonstrate that iteration time progressively decreases as additional quantization modules are incorporated. It also indicates that the training speed decreases from ZeRO-1 to ZeRO-2 and ZeRO-3, and our AGoQ significantly outperforms Megatron-LM, DeepSpeed, and ZeRO series.

\begin{table*}[!t]
\centering
\caption{Communication latency breakdown (ms) under 200Gbps and 10Gbps bandwidth.}
\label{tab:comm_savings}
\begin{tabular}{lccccc}
\toprule
Message Size & All-Reduce & All-to-All & Quant/Dequant & All-Gather & AGoQ \\
\midrule
$2^{30}$ (1GB) & 4292.77 / 50365.45 & 599.28 / 7297.73 & 31.03 / 31.07 & 556.47 / 226.58 & 1186.78 / 7555.38 \\
$2^{25}$ (32MB) & 131.23 / 1603.78 & 18.81 / 233.34 & 0.99 / 1.10 & 19.37 / 197.92 & 39.17 / 432.36 \\
$2^{20}$ (1MB) & 4.13 / 51.07 & 0.83 / 5.52 & 0.07 / 0.07 & 0.55 / 7.87 & 1.45 / 13.46 \\
$2^{15}$ (32KB) & 0.83 / 3.93 & 0.35 / 0.38 & 0.05 / 0.05 & 0.41 / 0.83 & 0.81 / 1.26 \\
\bottomrule
\end{tabular}
\end{table*}
\textbf{Communication Savings on Commodity Bandwidth (e.g., 100Gbps).} We evaluated communication efficiency under two representative bandwidth conditions: 200Gbps (our primary testbed) and 10Gbps (to simulate commodity constraints). Table~\ref{tab:comm_savings} reports latency breakdowns for both configurations (200Gbps / 10Gbps) under TP=8, DP=8. At 32MB under 200Gbps, our decomposed approach achieves a 3.4$\times$ speedup (39.17 ms vs. 131.23 ms). Under 10Gbps, the speedup increases to 3.7$\times$ (432.36 ms vs. 1603.78 ms), confirming that our method delivers substantial communication savings across diverse bandwidth settings.

\textbf{Wall-Clock Time Breakdown.} We performed a detailed timing breakdown for a single Transformer decoder layer (ms). Due to Megatron's sequence parallelism (SP), All-Gather costs are doubled during backward pass. Quantization/dequantization are fused into GEMM kernels. Table~\ref{tab:time_baseline} and Table~\ref{tab:time_agoq} report the breakdowns for the baseline and AGoQ, respectively. The results show that AGoQ introduces minimal overhead in compute-bound operations while effectively saving memory. The fused quantization/dequantization adds negligible latency, as evidenced by the small increase in FFN forward time (13.3 → 15.64 ms), FFN backward time (28.6 → 32.54 ms), Attn forward time (19.3 → 20.6 ms) and Attn backward time (45.1 → 47.8 ms).

\begin{table}[!h]
\centering
\caption{Baseline wall-clock time breakdown (ms) per Transformer decoder layer.}
\label{tab:time_baseline}
\adjustbox{max width=0.48\textwidth}{
\begin{tabular}{lcccccccc}
\toprule
Phase & ln & ag/rs & Attn & rs/ag & ln & ag/rs & FFN & rs/ag \\
\midrule
Forward & 1.3 & 17.02 & 19.34 & 21.79 & 1.3 & 17 & 13.3 & 23.01 \\
Backward & 2.85 & 23.21 & 45.06 & 35.4 & 3.2 & 22.62 & 28.6 & 34 \\
\bottomrule
\end{tabular}}
\end{table}

\begin{table}[!h]
\centering
\caption{AGoQ wall-clock time breakdown (ms) per Transformer decoder layer.}
\label{tab:time_agoq}
\adjustbox{max width=0.48\textwidth}{
\begin{tabular}{lcccccccc}
\toprule
Phase & ln & ag/rs & Attn & rs/ag & ln & ag/rs & FFN & rs/ag \\
\midrule
Forward & 1.3 & 16.82 & 20.62 & 21.65 & 1.3 & 17 & 15.64 & 21.78 \\
Backward & 2.9 & 22.82 & 47.86 & 34.56 & 3.4 & 22.58 & 32.54 & 33.54 \\
\bottomrule
\end{tabular}}
\end{table}
\textbf{Extended Throughput Analysis (2k/4k/8k).} We extended our throughput analysis to sequence lengths of 2k, 4k, and 8k. When memory is sufficient and recomputation is not required, we recommend enabling only gradient compression, as it effectively reduces communication overhead. Table~\ref{tab:throughput_ext} shows the throughput (samples/sec) comparison with Megatron-LM and ZeRO-1 baselines. Our method consistently outperforms both baselines across all sequence lengths, achieving speedups of up to 1.33$\times$ over Megatron-LM and 1.16$\times$ over ZeRO-1 at 2k sequence length.

\begin{table}[!h]
\centering
\caption{Throughput (samples/sec) comparison at sequence lengths 2k, 4k, and 8k.}
\label{tab:throughput_ext}
\begin{tabular}{lccc}
\toprule
Seq. & Megatron-LM & ZeRO-1 & AGoQ (Ours) \\
\midrule
2k & 2862.22 & 2498.13 & 2148.82 \\
4k & 4626.94 & 4261.65 & 3968.20 \\
8k & 8188.41 & 7848.12 & 7755.74 \\
\bottomrule
\end{tabular}
\end{table}

In the Appendix~\ref{subsec:ablation}, we also present several ablation studies.


\section{Conclusion}
In this work, we addressed the critical challenge of GPU memory consumption in LLM training through a holistic quantization approach.
Specifically, we present \modelname, which integrates: 1) a layer-aware activation quantization strategy that assigns suitable bits for storing activations based on layer types and pipeline parallelism stages, and 2) a gradient quantization algorithm that conserves memory and reduces communication time by using low-bit gradient storage and precision-preserved low-bit data All-Reduce communication.
Extensive experiments on two GPU clusters (up to 64 GPUs) demonstrate that \modelname{} reduces memory usage by 52\% compared to full-precision training and improves end-to-end training throughput up to 1.34$\times$ over state-of-the-art systems including Megatron-LM, DeepSpeed, ZeRO and COAT, while maintaining competitive accuracy on downstream tasks with LLaMA architectures. 

\bibliographystyle{icml2026}
\bibliography{ref}

\clearpage
\section{Appendix}
\subsection{Error Analysis of Activation Quantization}\label{ap:error_analy}
We conduct an error analysis for SiLU \& Multiply, RMSNorm+GEMM, and Attention mentioned in \S~\ref{sec:error_ana}.
\subsubsection{SiLU \& Multiply}
SiLU \& Multiply is an element-wise operation that takes two inputs, denoted as $X$ and $Y$. For a pair of corresponding elements $x$ and $y$ from these inputs, the SiLU \& Multiply operation is defined as:
\[
z = x y \sigma(y)
\]
where $\sigma$ is the sigmoid function.
Its derivatives are
\[
\frac{\partial z}{\partial x} = y\sigma(y), \quad \frac{\partial z}{\partial y} = x\sigma(y) + x y \sigma(y)(1 - \sigma(y)),
\]
and perturbations are
\[
x' = x(1 + \delta x), \quad y' = y(1 + \delta y).
\]

\textbf{Case 1 (Recompute intermediate values)}

\[
\begin{aligned}
\Delta \frac{\partial z}{\partial x} &\approx (\sigma(y) + y\sigma'(y))y\delta y.
\end{aligned}
\]
\[
\begin{aligned}
|\Delta \frac{\partial z}{\partial x}| &\leq |y|(\sigma(y) + |y|\sigma(y)(1-\sigma(y)))|\delta y| \\
&\leq \mathcal{O}(|y|^2|\delta y|).
\end{aligned}
\]
For \( \partial z/\partial y \):
\[
\begin{aligned}
\Delta \frac{\partial z}{\partial y} \approx & x\sigma(y)(1+y(1-\sigma(y)))\delta x \\
&+ x\sigma(y)(1-\sigma(y))(2+y(1-2\sigma(y)))y\delta y.
\end{aligned}
\]
\[
\begin{aligned}
|\Delta \frac{\partial z}{\partial y}| \leq & |x|\sigma(y)|1+y(1-\sigma(y))||\delta x| \\
&+ |x|\sigma(y)(1-\sigma(y))|y||2+y(1-2\sigma(y))||\delta y|.
\end{aligned}
\]
Asymptotically: \( \mathcal{O}(|x||\delta x| + |x||y||\delta y|) \).

\textbf{Case 2 (Cache intermediate values)}
\[
\sigma' \approx \sigma(y)(1+\delta s), \qquad 
\frac{\partial z'}{\partial x'} = y'\sigma' \approx y(1+\delta y)\sigma(y)(1+\delta s).
\]
\[
\Delta \frac{\partial z}{\partial x} \approx y\sigma(y)(\delta y + \delta s).
\]
\[
|\Delta \frac{\partial z}{\partial x}| \le |y|\sigma(y)(|\delta y| + |\delta s|) \le \mathcal{O}\big(|y|(|\delta y|+|\delta s|)\big).
\]
For \(\frac{\partial z'}{\partial y'}\):
\[
\frac{\partial z'}{\partial y'} = x'\sigma' + x'y'\sigma'(1-\sigma').
\]
Expanding to first order:
\[
\begin{aligned}
\frac{\partial z'}{\partial y'} &\approx x\sigma(y)\big[1 + \delta x + \delta s\big] \\
&+ x y \sigma(y)(1-\sigma(y))\big[1 + \delta x + \delta y + \delta s - \tfrac{\sigma(y)}{1-\sigma(y)}\delta s\big].
\end{aligned}
\]
Subtracting the exact \(\frac{\partial z}{\partial y} = x\sigma(y) + x y \sigma(y)(1-\sigma(y))\):
\begin{align*}
\Delta \frac{\partial z}{\partial y} &\approx 
x\sigma(y)\big[1 + y(1-\sigma(y))\big]\delta x \\
&\quad + x y \sigma(y)(1-\sigma(y))\delta y \\
&\quad + x\sigma(y)\big[1 + y(1-2\sigma(y))\big]\delta s.
\end{align*}
\begin{align*}
\left|\Delta \frac{\partial z}{\partial y}\right| &\le 
|x|\sigma(y)\big|1 + y(1-\sigma(y))\big||\delta x| \\
&\quad + |x||y|\sigma(y)(1-\sigma(y))|\delta y| \\
&\quad + |x|\sigma(y)\big|1 + y(1-2\sigma(y))\big||\delta s|.
\end{align*}
Asymptotically:  
\[
|\Delta \frac{\partial z}{\partial y}| = \mathcal{O}\big(|x|(|\delta x|+|\delta y| + |y||\delta s|)\big).
\]

Comparing the two cases for the SiLU \& Multiply operation, under the common scenario where the inputs \(x\) and \(y\) are mostly smaller than 1, **Case 1 (recomputing intermediate values) gives a strictly smaller error upper bound** than Case 2 (caching intermediate values).  
For \(\partial z/\partial x\), Case~1 yields \(\mathcal{O}(|y|^2|\delta y|)\) while Case~2 yields \(\mathcal{O}(|y|(|\delta y|+|\delta s|))\); because \(|y|\le 1\) implies \(|y|^2|\delta y| \le |y||\delta y| \le |y|(|\delta y|+|\delta s|)\), the recompute bound is always tighter.  
For \(\partial z/\partial y\), Case~1's asymptotic bound \(\mathcal{O}(|x||\delta x| + |x||y||\delta y|)\) is also lower than Case~2's \(\mathcal{O}(|x|(|\delta x|+|\delta y| + |y||\delta s|))\), since the latter contains an extra term proportional to \(|\delta s|\) that is absent when the sigmoid is recomputed exactly from the perturbed input.  
Thus, as long as the cache error \(\delta s\) is non‑negligible and the typical input magnitudes satisfy \(|x|,|y| < 1\), recomputing the intermediate values on the fly produces a provably smaller worst‑case gradient error.
\textbf{Therefore, for SiLU computation, we only store the quantized input activations and recompute the necessary intermediate values during the backward pass.}

\subsubsection{RMSNorm + GEMM}
Consider \( Y = W U \), where \( U = \text{RMSNorm}(X) \). The gradient w.r.t. \( W \) is
\[
\frac{\partial Y}{\partial W} = U^T, \quad U = \gamma X / r.
\]

\textbf{Case 1 (Recompute intermediate values)}

Input perturbation:
\[
X' = X\odot (1 + \delta X), \quad r' = r(X'), \quad U' = \gamma X' / r'.
\]
Gradient error:
\[
\Delta \left( \frac{\partial Y}{\partial W} \right) = U'^T - U^T.
\]
First-order expansion:
\[
\|U' - U\|_2 \le  \|\gamma\|_\infty \left( \frac{\|X\|_2}{r^2} + \frac{\|X\|_2^3}{d r^4} \right) \|\delta X\|_\infty.
\]
Asymptotically: \( \mathcal{O}(d\| \gamma\|_\infty\|\delta X\|_\infty / \|X\|_2)\).

\textbf{Case 2 (Cache intermediate values)}
\[
U_c = U\odot(1+ \delta U).
\]
Gradient error bound:
\[
\|\Delta U\|_2 = \|U \odot\delta U\|_2 \le \|U\|_2 \|\delta U\|_\infty.
\]
Asymptotically: \( \mathcal{O}(\|U\|_2 \|\delta U\|_\infty) \).

Two cases yield the same asymptotic error order, with only constant-factor differences. \textbf{Given that recompute also reduces memory footprint, for GEMM computation, we only keep the quantized input activations of RMSNorm.} Similarly, it was found that storing only the input activations of SiLU and recalculating the input of GEMM during gradient computation for GEMM after SiLU also decreases the upper bound on gradient error.

\subsubsection{Attention}
A single-head attention is:
\[
A = PV = \text{softmax}(S)V, \quad S = \frac{QK^T}{\sqrt{d}}.
\]
Its gradients are 
\[
\begin{aligned} 
&\frac{\partial A}{\partial V} = P^\top , & &\frac{\partial A}{\partial P} =  V^\top, \\
&\frac{\partial A}{\partial S} = P    (\frac{\partial A}{\partial P} - z \mathbf{1}^\top), & &z = \text{rowsum}(P    \frac{\partial A}{\partial P}), \\
&\frac{\partial A}{\partial Q} = \frac{1}{\sqrt{d}} \frac{\partial A}{\partial S} \, K, & &\frac{\partial A}{\partial K} = \frac{1}{\sqrt{d}} (\frac{\partial A}{\partial S})^\top Q,
\end{aligned}
\]
and perturbations are
\[
Q' = Q\odot(1 + \delta Q), \quad
K' = K\odot(1 + \delta K), \quad
V' = V\odot(1 + \delta V). \\
\]
\textbf{Case 1 (Recompute intermediate values).}
\[
A' = \text{attention}(Q',K',V')
\]
Thus, the gradient errors can be represented as: 
\begin{align*}
\|\Delta A\|_2 &\leq \|\Delta S\|_2 \|V\|_2 + \|V\odot \delta V\|_2 \\
\|\Delta \frac{\partial A}{\partial V}\|_2 &\leq \frac{1}{\sqrt{d}} \|Q\odot \delta Q K^\top + Q(K\odot\delta K)^\top\|_2 \\
&= \mathcal{O}\left( \frac{\|Q\|_2 \|K\|_2}{\sqrt{d}} (\|\delta Q\|_\infty + \|\delta K\|_\infty) \right).
\end{align*}

\begin{align*}
\|\Delta \frac{\partial A}{\partial K}\|_2 &\leq \frac{\|V\|_2 \|K\|_2}{d} \|Q\odot \delta Q K^\top + Q(K\odot\delta K)^\top\|_2 \\
&= \mathcal{O}\left( \frac{\|Q\|_2 \|K\|_2 \|V\|_2}{d} (\|\delta Q\|_\infty + \|\delta K\|_\infty) \right).
\end{align*}

\begin{align*}
\|\Delta \frac{\partial A}{\partial Q}\|_2 &\leq \frac{\|V\|_2 \|Q\|_2}{d} \|Q\odot\delta Q K^\top + Q(K\odot\delta K)^\top\| _2\\
&= \mathcal{O}\left( \frac{\|V\|_2 \|Q\|_2 \|K\|_2}{d} (\|\delta Q\|_\infty + \|\delta K\|_\infty) \right).
\end{align*}

\textbf{Case 2 (Cache intermediate values).}

Since
\[
A' =A\odot(1+\delta A),
\]
the gradient errors in this case are:
\begin{align*}
\|\Delta \frac{\partial A}{\partial V}\|_2 &\leq \frac{\|A\|_2}{\|V\|_2} \left( \frac{1}{\sqrt{d}} \|Q\odot\delta Q K^\top \right. \\
&\quad \left. + Q(K\odot\delta K)^\top\|_2 + \|A\odot\delta A\| _2\right) \\
&= \mathcal{O}\left( \frac{\|Q\|_2 \|K\|_2}{\sqrt{d}} (\|\delta Q\|_\infty + \|\delta K\|_\infty) \right).
\end{align*}

\begin{align*}
\|\Delta \frac{\partial A}{\partial K}\| _2&\leq \frac{2}{\sqrt{d}} \|V\|_2 \|Q\odot\delta Q K^\top + Q(K\odot\delta K)^\top\|_2 \\
&\quad + 2\|V\|_2\|\delta V\|_\infty + \|A\odot\delta A\|_2 \\
&= \mathcal{O}\left( \frac{\|Q\|_2 \|K\|^2 _2\|V\|_2}{d} (\|\delta Q\|_\infty + \|\delta K\|_\infty) \right).
\end{align*}

\begin{align*}
\|\Delta \frac{\partial A}{\partial Q}\| _2&\leq \frac{1}{\sqrt{d}} \left[ \frac{2}{\sqrt{d}} \|V\|_2 \|K\| _2\|Q\odot\delta Q K^\top \right. \\
&\quad + Q(K\odot\delta K)^\top\| _2+ 2\|V\| _2\|K\| _2\|\delta V\|_\infty \\
&\quad \left. + \|K\|_2 \|A\odot\delta A\|_2 + \|V\| _2\|\delta K\|_\infty \right] \\
&= \mathcal{O}\left( \frac{\|V\|_2 \|Q\|^2 _2\|K\|_2}{d} (\|\delta Q\|_\infty + \|\delta K\|_\infty) \right).
\end{align*}
Based on the derived error bounds under the “recompute intermediate values” case (Case~1), we compare the gradient perturbations of RMSNorm and the attention operation. According to Eq.~\ref{equ:rmsnorm-case1-error} and the subsequent analysis, the RMSNorm gradient error satisfies \(\|\Delta J\|_2 = \mathcal{O}(\eta)\) under standard Transformer assumptions (\(\|X\|_2^2 = \Theta(d)\), \(r = \Theta(1)\), and \(\|\gamma\|_\infty = \Theta(1)\)), where \(\eta = \|\delta X\|_\infty\) is the normalized input perturbation level. In contrast, the attention gradient errors scale much more severely with the sequence length \(L\) and the per-head dimension \(d_k\). For \(\partial A / \partial V\) we obtain \(\|\Delta (\partial A/\partial V)\|_2 = \mathcal{O}(\eta L \sqrt{d_k})\), while for \(\partial A/\partial K\) and \(\partial A/\partial Q\) the bounds grow to \(\mathcal{O}(\eta L^{3/2} \sqrt{d_k})\). These quantities are larger than the RMSNorm error by factors of \(\Theta(L\sqrt{d_k})\) and \(\Theta(L^{3/2}\sqrt{d_k})\), respectively. Such a multiplicative gap explains why activation quantization applied to the Q, K, V projections leads to severely amplified gradient errors and training instability, whereas RMSNorm can be safely quantized. \textbf{Therefore, for the attention operation, we choose not to apply activation quantization.}

\begin{table}[!t]
\centering
\caption{GPU and NPU Memory Usage (GB) Comparison}
\label{tab:quant_comparison}
\begin{tabular}{lcccc}
\toprule
GPU/NPU & \textbf{AGoQ} & \textbf{O+G} & \textbf{O} & \textbf{Megatron-LM} \\
\midrule
GPU & 22.3 & 35.3 & 37.7 & 46.1 \\
NPU & 29.7 & 40.5 & 45.3 & 55.2 \\
\bottomrule
\end{tabular}
\end{table}

\begin{figure}[!t]
	\centering
 	\includegraphics[width=0.48\textwidth]{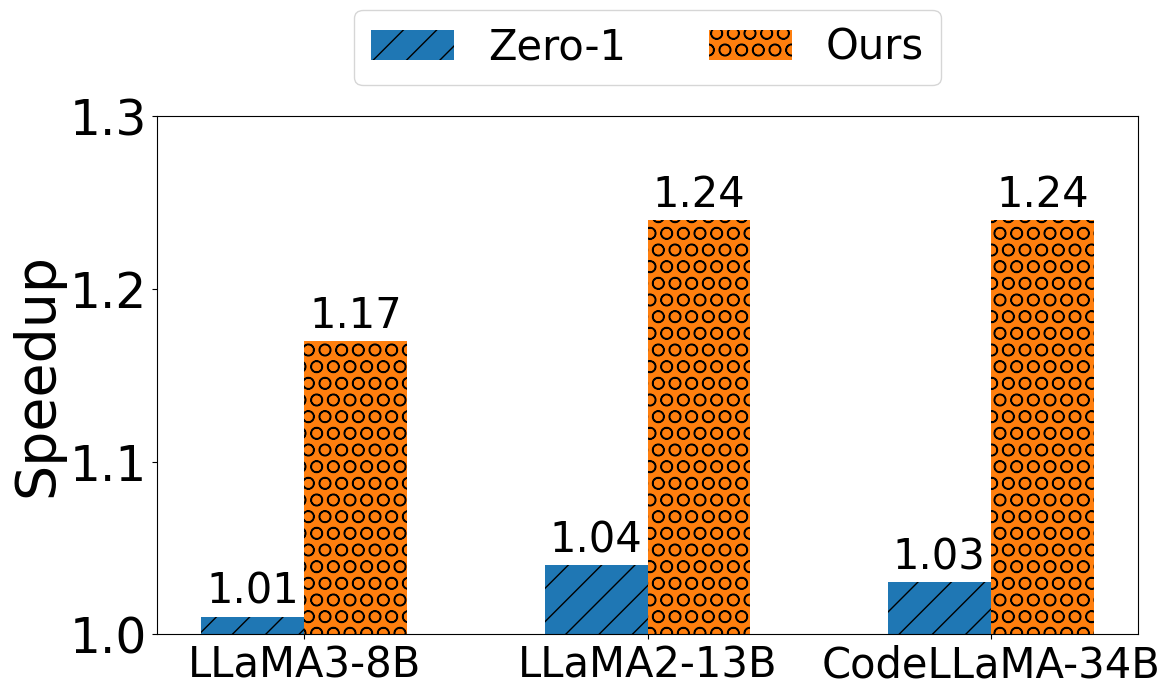}
	\caption{Speedups of our \modelname{} and ZeRO-1 over Megatron-LM on varied models.}
	\label{fig:models}
\end{figure}

\begin{table}[h]
\centering
\small
\caption{Experimental configurations.}
\label{tab:exp_config}
\begin{tabular}{@{}ccccc@{}}
\toprule
\textbf{\# of GPUs} & \textbf{Model} & \textbf{Seq. Len.} & \textbf{TP} & \textbf{PP} \\
\midrule
8 & LLaMA3-8B & 16K--32K & 4 & 1 \\
8 & LLaMA3-8B & 32K--80K & 8 & 1 \\
16 & LLaMA2-13B & 32K--80K & 8 & [1,2] \\
32 & LLaMA2-13B & 32K--80K & 8 & [1,2,4] \\
32 & CodeLLaMA-34B & 32K--80K & 8 & [1,2,4] \\
64 & LLaMA2-13B & 32K--80K & 8 & [1,2,4,8] \\
64 & CodeLLaMA-34B & 32K--80K & 8 & [1,2,4,8] \\
\bottomrule
\end{tabular}
\end{table}



\begin{table}[!t]
\centering
\caption{Accuracy changes ($\Delta$ Acc) of AGoQ compared to the Megatron-LM FP16 baseline across datasets. Results are shown for LLaMA2-7B (2B tokens) and LLaMA3.2-1B (10B tokens).}
\label{tab:accuracy_changes}
\adjustbox{max width=0.48\textwidth}{
\begin{tabular}{l|ccc|ccc}
\toprule
\multirow{2}{*}{Dataset} & \multicolumn{3}{c|}{LLaMA2-7B (2B tokens)} & \multicolumn{3}{c}{LLaMA3.2-1B (10B tokens)} \\
\cmidrule(lr){2-4} \cmidrule(lr){5-7}
& Baseline & AGoQ & $\Delta$ & Baseline & AGoQ & $\Delta$ \\
\midrule
arc\_c   & 0.1988 & 0.1834 & -0.0154 & 0.1877 & 0.2099 & +0.0222 \\
arc\_e   & 0.4179 & 0.4158 & -0.0021 & 0.4571 & 0.4714 & +0.0143 \\
hellas.  & 0.2886 & 0.2897 & +0.0011 & 0.3276 & 0.3298 & +0.0022 \\
piqa     & 0.5990 & 0.6039 & +0.0049 & 0.6219 & 0.6284 & +0.0065 \\
sciq     & 0.7260 & 0.7280 & +0.0020 & 0.7180 & 0.7100 & -0.0080 \\
winog.   & 0.4830 & 0.5036 & +0.0206 & 0.5193 & 0.5185 & -0.0008 \\
\bottomrule
\end{tabular}
}
\end{table}

\subsection{Ablation Studies}\label{subsec:ablation}
\textbf{Performance on Varied Models.}
To evaluate improvements in our methods across different models, we organize the experiments according to the setup in Table~\ref{tab:exp_config}. Specifically, we compare the speedup of Zero-1 and \modelname{} against Megatron-LM on LLaMA3-8B, LLaMA2-13B, and CodeLLaMA-34B, as shown in Fig.~\ref{fig:models}. The results demonstrate that our method consistently delivers substantial gains across all models. Detail configurations are shown in Table.~\ref{tab:exp_config}.

\textbf{Zero-shot Accuracy.}
We further assess zero-shot accuracy of LLaMA2-7B trained on 2B tokens and LLaMA3.2-1B tained on 10B tokens using ARC-Challenge~\cite{arc_hel}, ARC-Easy~\cite{arc_hel}, HellaSwag~\cite{arc_hel}, PIQA~\cite{bisk2020piqa}, SciQ~\cite{sciq}, and Winogrande~\cite{sakaguchi2021winogrande}. Table~\ref{tab:accuracy_changes} shows that the mean accuracy across all six datasets remains high with no degradation. 

\textbf{Memory Footprint Reduction.} We also compare the effects of activation quantization, gradient quantization, and optimizer quantization on memory reduction in Table~\ref{tab:quant_comparison} with TP=8, PP=1, sequence length of 12K on LLaMA2-13B with GPUs and NPUs. As can be seen from the table, each quantization module contributes significantly to reducing memory usage. Specifically, only applying 8-bit optimizer quantization (``O'')~\cite{dettmers2022bit}, it saves around 8G memory size on GPU and 10G on NPU, while our activation and gradient quantization further save 13G and 2.4G memory on GPU while 10.8G and 4.8G on NPU, respectively. Notably, our \modelname{} (``A+O+G'') achieves a $53\%$ reduction of the peak memory footprint over Megatron-LM on GPU while $46\%$ on NPU.

\begin{table}[!t]
\centering
\caption{Speedups of kernel fusion of GEMM and quantization/dequantization over sequential operations.}
\label{tab:matrix_speedup}
\begin{tabular}{ccccr}
\toprule
\textbf{$m$} & \textbf{$k$} & \textbf{$n$} & \textbf{Speedup} \\
\midrule
16,384 & 14,336 & 4,096 & 1.08x \\
16,384 & 4,096  & 14,336 & 1.03x \\
16,384 & 4,096  & 12,288 & 1.04x \\
16,384 & 4,096  & 4,096 & 1.11x \\
\bottomrule
\end{tabular}
\end{table}

\textbf{Improvement of kernel fusion.} 
We test the speedup achieved by kernel fusion (\S\ref{subsec:kernelfusion}) of GEMM computation with quantization or dequantization operations. As shown in Table~\ref{tab:matrix_speedup}, we demonstrate several major matrix shapes for GEMM from LLaMA3-8B as examples, where $m, k, n$ represent the dimensions of GEMM $C=A\times B$, where $A\in\mathbb{R}^{m\times k}$ and $B\in \mathbb{R}^{k\times n}$. The results demonstrate that our kernel fusion approach achieves an average speedup of 1.07$\times$ over the sequential version of GEMM and quantization/dequantization, which results in an end-to-end speedup of 1.05$\times$ on LLaMA3-8B with PP=1, TP=8 and sequence length of 16K.

\textbf{Ablation for DBC (Dynamic Block Compression).} We conducted an ablation study comparing training with and without DBC. Table~\ref{tab:dbc_ablation} shows that DBC consistently improves accuracy on most tasks. Notably, DBC configured under low pipeline parallelism (PP) can be applied to high PP without increasing peak memory, highlighting its flexibility.

\begin{table}[!h]
\centering
\caption{Ablation study of DBC: accuracy (\%) with and without DBC on six datasets.}
\label{tab:dbc_ablation}
\adjustbox{max width=0.48\textwidth}{
\begin{tabular}{lcccccc}
\toprule
 & arc\_c & arc\_e & hellas. & piqa & sciq & winog. \\
\midrule
w/o DBC & 17.66 & 41.88 & 28.53 & 60.12 & 71.20 & 50.83 \\
w/ DBC & 18.34 & 41.58 & 28.97 & 60.39 & 72.80 & 50.36 \\
\bottomrule
\end{tabular}}
\end{table}

\textbf{Ablation for Layer-Aware Bit Width.} We compared our layer-aware mixed-precision quantization against a naïve uniform 4-bit quantization applied to all activations. The uniform 4-bit baseline failed to converge, underscoring the importance of adaptive precision.

\textbf{Actual Per-Layer Gradient Error.} We measured the gradient error of input activations for each layer (mean absolute error and normalized L2 distance) introduced by AGoQ compared to full-precision gradients, sampled from a model trained after 10B tokens. For GEMM, we additionally compare the gradient error of the weight. Table~\ref{tab:gradient_error} reports the results. The normalized L2 errors for Attention (0.14–0.15) are the largest among all layers, which aligns with our theoretical analysis and explains why we exclude Attention from quantization. For other layers, the minimum normalized error is 0.003 (LayerNorm) and the maximum is 0.051 (SiLU), both of which have negligible impact on convergence.

\begin{table}[!h]
\centering
\caption{Per-layer gradient error (mean absolute error and normalized L2 distance) introduced by AGoQ.}
\label{tab:gradient_error}
\begin{tabular}{lcc}
\toprule
Layer & MAE & Normalized L2 \\
\midrule
LayerNorm & $2.8 \times 10^{-10}$ & 0.003 \\
GEMM (Weight) & $5.3 \times 10^{-7}$ & 0.026 \\
SiLU & $2.6 \times 10^{-9}$ & 0.051 \\
Attention Q & $2.5 \times 10^{-9}$ & 0.14 \\
Attention K & $1.7 \times 10^{-9}$ & 0.15 \\
Attention V & $2.0 \times 10^{-9}$ & 0.059 \\
\bottomrule
\end{tabular}
\end{table}

\textbf{Memory Reduction at 32k/64k Sequence Lengths.} We conducted memory usage experiments on Llama3-8B with extended sequence lengths under memory constraints. For 32k sequence length, we configured 36 layers; for 64k, we used 16 layers due to memory limitations. As shown in Table~\ref{tab:memory_long}, AGoQ achieves substantial memory savings, reducing footprint by up to 66\% at 32k and 59\% at 64k, demonstrating its effectiveness in memory-constrained long-sequence scenarios.

\begin{table}[!h]
\centering
\caption{Memory consumption (MB) for Llama3-8B at 32k and 64k sequence lengths.}
\label{tab:memory_long}
\begin{tabular}{lcc}
\toprule
Sequence Length & Baseline (MB) & AGoQ (MB) \\
\midrule
32k (36 layers) & 48606 & 16594 \\
64k (16 layers) & 46681 & 19267 \\
\bottomrule
\end{tabular}
\end{table}

\textbf{Gradient Norm Monitoring.} We conducted an extended training run on Llama3-8B with a per-iteration token budget of 32k, spanning 100,000 iterations. Table~\ref{tab:gradient_norm} reports average gradient norms at 10k-iteration intervals. AGoQ maintains gradient norms closely aligned with the baseline throughout training, with stable convergence.

\begin{table}[!h]
\centering
\caption{Gradient norm comparison every 10k iterations over 100k iterations.}
\label{tab:gradient_norm}
\begin{tabular}{lcc}
\toprule
Iteration Range & AGoQ & Baseline \\
\midrule
0–10k & 7.28 & 7.10 \\
10k–20k & 4.33 & 4.38 \\
20k–30k & 3.79 & 3.77 \\
30k–40k & 3.64 & 3.36 \\
40k–50k & 3.38 & 3.20 \\
50k–60k & 3.32 & 3.10 \\
60k–70k & 3.17 & 3.15 \\
70k–80k & 3.32 & 3.29 \\
80k–90k & 3.20 & 3.37 \\
90k–100k & 3.08 & 3.21 \\
\bottomrule
\end{tabular}
\end{table}

\end{document}